\documentclass[conference]{IEEEtran}
\IEEEoverridecommandlockouts
\usepackage{cite}
\usepackage{amsmath,amssymb,amsfonts}
\usepackage{algorithmic}
\usepackage{graphicx}
\usepackage{textcomp}
\usepackage{xcolor}
\def\BibTeX{{\rm B\kern-.05em{\sc i\kern-.025em b}\kern-.08em
    T\kern-.1667em\lower.7ex\hbox{E}\kern-.125emX}}

\usepackage{microtype}  
\renewcommand{\baselinestretch}{0.96}
\begin{document}

\title{Towards Wearable Interfaces for Robotic Caregiving \\
\thanks{This research was supported by the National Science Foundation Graduate Research Fellowship Program under Grant No. DGE1745016 and DGE2140739.}
}

\author{\IEEEauthorblockN{Akhil Padmanabha}
\IEEEauthorblockA{\textit{Robotics Institute} \\
\textit{Carnegie Mellon University}\\
akhilpad@andrew.cmu.edu}
\and
\IEEEauthorblockN{Carmel Majidi}
\IEEEauthorblockA{\textit{Department of Mechanical Engineering} \\
\textit{Carnegie Mellon University}\\
cmajidi@andrew.cmu.edu}
\and
\IEEEauthorblockN{Zackory Erickson}
\IEEEauthorblockA{\textit{Robotics Institute} \\
\textit{Carnegie Mellon University}\\
zackory@cmu.edu}
}

\maketitle
\begin{abstract}
Physically assistive robots in home environments can enhance the autonomy of individuals with impairments, allowing them to regain the ability to conduct self-care and household tasks. Individuals with physical limitations may find existing interfaces challenging to use, highlighting the need for novel interfaces that can effectively support them. In this work, we present insights on the design and evaluation of an active control wearable interface named HAT, Head-Worn Assistive Teleoperation. To tackle challenges in user workload while using such interfaces, we propose and evaluate a shared control algorithm named Driver Assistance. Finally, we introduce the concept of passive control, in which wearable interfaces detect implicit human signals to inform and guide robotic actions during caregiving tasks, with the aim of reducing user workload while potentially preserving the feeling of control.
\end{abstract}


\section{Introduction}

Caregiving robots in the home can assist individuals with impairments in performing physical tasks, enhancing their autonomy and quality of life.~\cite{tasks, tasksaroundhead, yang2023high, madan2022sparcs, nanavati2023physically, padmanabha2024voicepilot, yuan2024towards}. The development and evaluation of novel assistive interfaces for caregiving robots could lead to alternatives for individuals with impairments who may have a difficult time using conventional options. Wearables using sensing modalities such as accelerometers, inertial measurement units (IMUs), and contact microphones have been useful for the identification of various human movements~\cite{heikenfeld2018wearable, yetisen2018wearables, padmanabha2023multimodal}. Interfaces incorporating these sensing modalities could enable individuals with motor impairments to more effectively control mobile manipulators~\cite{padmanabha2023hat, padmanabha2024independence, yang2023high}. 

In this work, we present insights from designing wearable interfaces for active and shared control of caregiving robots, while also introducing the concept of passive control wearable interfaces. Active control refers to situations where the user's intentional actions or movements directly trigger robotic movements. This includes direct teleoperation interfaces, such as web-based controls~\cite{robotsforhumanity, ranganeni2024customizing}, or physical buttons that manually trigger sequences of robot actions like having a robot deliver a spoonful of food in robot-assisted feeding~\cite{Obi}. While active control offers the user the highest degree of control, it often results in higher user workload. To address this, shared control can combine user input from direct teleoperation interfaces with the robot's autonomy, leveraging the robot's understanding of human intentions and the surrounding environment~\cite{selvaggio2021autonomy, brenna2018autonomy, hameed2023control} and resulting in a reduction in cognitive load for the user~\cite{merkt2017robust, losey2018review, gopinath2017human}. However, increased autonomous assistance often comes at the cost of the user's sense of control and agency~\cite{collier2025sense, javdani2018shared, bhattacharjee2020more}. 

To further minimize workload while preserving the sense of user control, we introduce a new control paradigm for wearable interfaces: passive control. Passive control involves a robot detecting and acting directly on implicit signals from wearable sensors. We draw motivation for passive control from human caregivers who often rely on cues from care recipients during assistive tasks. For example, during feeding assistance, a care recipient may implicitly signal readiness for the next bite by finishing chewing or swallowing, along with cues like eye contact or turning their head toward the caregiver. Passive control could offer a method to further minimize workload while maintaining the feeling of control by sensing subtle, subconscious human movements, which the robot can interpret to initiate or modify its actions. 

The research questions of interest are:
\begin{itemize}
    \item \textbf{RQ1} What benefits and drawbacks do wearable robotic interfaces have for the active control of physically assistive mobile manipulators?
    \item \textbf{RQ2:} How can we reduce user workload in wearable teleoperation interfaces through shared control? 
    \item \textbf{RQ3:} How can we further reduce workload and maintain the feeling of control by using wearable sensing methods for passive control of physically assistive robots?
\end{itemize}

\begin{figure*}[hbt!]
      \centering
      \includegraphics[width = \textwidth]{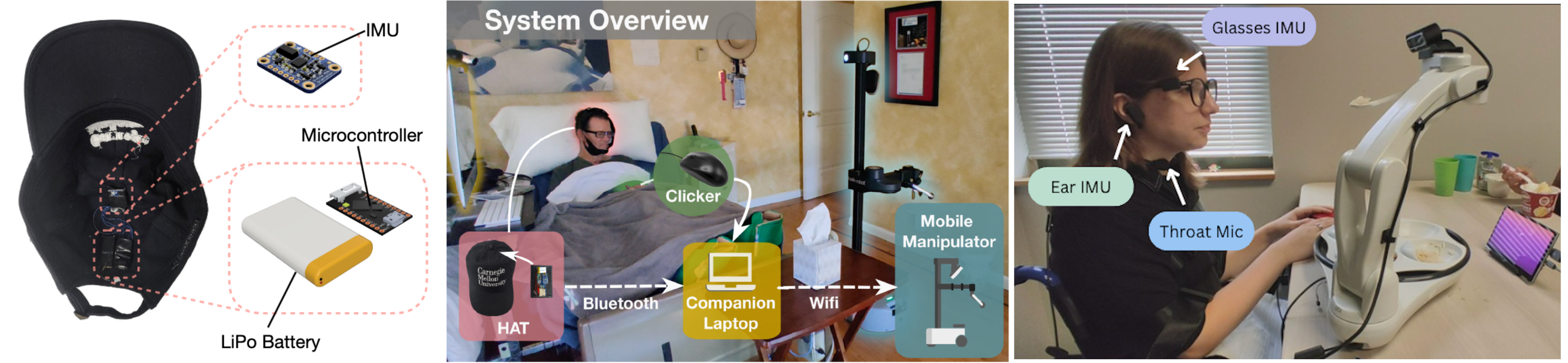}
      \caption{Left: The wireless head-worn interface (HAT) with integrated inertial measurement unit (IMU) sensing. Middle: Henry Evans, a non-speaking individual with quadriplegia, uses HAT to teleoperate a mobile manipulator. The interface records and wirelessly sends head orientation angles to a companion laptop, which computes and transmits actuator velocities to the robot. A clicker is used to switch between modes, distinct operational states of the HAT system. Right: Our system for passive control using wearable sensing for robot assisted feeding is shown. The wearable sensors include glasses and ear IMUs and a throat contact microphone.}
      \label{fig:theonefigure}  
      \vspace{-0.5cm}
   \end{figure*}

\section{RQ1: Active Control}
\label{sec:RQ1}
To investigate RQ1, we designed a wearable interface named HAT (Head-Worn Assistive Teleoperation)~\cite{padmanabha2023hat, padmanabha2024independence} for control of a high degree of freedom mobile manipulator, the Hello Robot Stretch~\cite{stretch}. Developing wearable interfaces presents unique challenges, including ensuring precise sensing and intuitive mapping of human movements to robot actions, minimizing latency, and accommodating the needs of users with varying levels of impairments. We explore these challenges using HAT, shown in Fig.~\ref{fig:theonefigure}, which serves as a direct teleoperation interface that maps head orientation angles from an inertial measurement unit (IMU) to actuator velocities. We initially evaluated HAT in 2 hour studies in a lab environment with both non-disabled and individuals with impairments~\cite{padmanabha2023hat}\footnote{All studies detailed in this work were approved by Carnegie Mellon University's IRB.}. We subsequently iterated on the interface and system design using both quantitative and qualitative data from the initial study in conjunction with input from Henry Evans, a non-speaking individual with quadriplegia who has participated extensively in assistive robotics studies. This process culminated in a 7 day evaluation study with Henry in his home setting, shown in Fig.~\ref{fig:theonefigure}~\cite{padmanabha2024independence}. 

In response to RQ1 about identifying the benefits and drawbacks of active control wearable interfaces, through our studies, we find that such interfaces offer distinct advantages when compared to standard interfaces for robotic teleoperation. First, wearable interfaces embedded in clothes are readily accessible when needed and inconspicuous when not in use. Second, wearable interfaces can offer a more direct and intuitive means of controlling assistive robots. The absence of a screen or input device (head tracking, sip and puff, etc.) directly in front of the user's face allows for increased flexibility when performing tasks around the body and enhances situational awareness. However, we also find through our studies that wearable interfaces used for direct teleoperation struggle from high user workload similar to other direct teleoperation interfaces. We address these challenges in RQ2 and RQ3. 

\section{RQ2: Shared Control}
\label{sec:RQ2}
To minimize user workload, initially identified in our first study with HAT, we propose a shared control (SC) method for HAT and other teleoperation interfaces called Driver Assistance (DA). When DA is activated by the user, the system uses an open-vocabulary object detection perception model, OWL-ViT~\cite{minderer2205simple}, to process input language queries that define target objects and match these to detected objects within the environment. It then automatically aligns the robotic gripper's position with the intended object, While developing SC methods, the user's feeling of control over the system is an important consideration as previous studies show that higher levels of autonomy can lead to a reduced feeling of control~\cite{javdani2018,bhattacharjee2020, collier2025sense}. Our proposed method limits the shared control to specific robot joints to ensure that the user maintains a sense of control.

We directly compared our shared control method against active control direct teleoperation with Henry Evans in the aforementioned 7-day in-home study across 3 caregiving tasks. We find that DA led to clear improvements in task times and workload measures while still preserving user control of the robot in the home. For fetching a Red Bull can, DA reduced the task time by 70\%. DA also reduced errors during grasping tasks, lessened the user's dependence on clear object perception using line of sight or from the robot's camera views, and led to a 4 point reduction in both mental demand and effort using a 7-point NASA TLX Scale. While further testing needs to be conducted with more users, we believe these preliminary findings show that incorporating open-vocabulary object detection perception models alongside shared control for alignment during grasping tasks may play a vital role in alleviating the workload and perception demands involved in robotic teleoperation using wearable interfaces.

\section{RQ3: Passive Control}
\label{sec:RQ3}
For RQ3, we investigate how passive control wearable interfaces equipped with motion sensing, such as HAT, can enable robots to detect and act on implicit user signals, reducing input demands while preserving user control through intuitive interactions. We commence our work with robot-assisted feeding, addressing the specific challenge of determining bite timing, the moment when the robot should feed the care recipient. Previous studies have relied on the user manually triggering bite timing~\cite{Obi, gordon2024adaptable} or have been limited to specific contexts, such as social dining settings~\cite{ondras2022human}. 

We present a system, shown in Fig.~\ref{fig:theonefigure}, consisting of wearable sensors, including head-mounted IMUs and a contact microphone on the user's neck. These sensors provide our method with cues often used by caregivers during feeding including the user's head pose and when they are chewing, swallowing, and talking. Preliminary testing has shown that all the aforementioned cues are visible in the raw sensor data. We conducted a human study with non-impaired participants with half the participants doing a participant-controlled version of the study, while the other half participate in a Wizard of Oz version. We employ the participant-controlled version to collect well-labeled bite timing data, while the Wizard of Oz condition gathers data on the participant's interactions with an autonomous robot, including their reactions to errors in bite timing. We are in the process of training data-driven machine learning algorithms to estimate bite timing using our wearable sensors. Following this, we will run evaluation studies with both non-disabled individuals and those with motor impairments to compare our passive control wearable approach to active control methods such as manual triggering. We anticipate this work will reveal unique insights, as our approach differs from most prior research on wearable interfaces by removing the requirement for continuous active input, which can be demanding for many users with impairments. Our methods for passive control using wearables could also be extended past robot assisted feeding to other tasks in robotic caregiving such as robot assisted dressing.

\bibliographystyle{IEEEtran}
\bibliography{bibliography}

\end{document}